\documentclass[10pt,journal,twocolumn]{IEEEtran}
 \usepackage[utf8x]{inputenc}
\usepackage{amssymb}
\usepackage{graphicx}
\usepackage{multicol}
\usepackage{balance}
\usepackage{stfloats}
\usepackage[cmex10]{amsmath}
\usepackage{amsmath}

\newtheorem{theo}{\bf Theorem}[section]

\begin{document}

\title{Patch-Based  Low-Rank Minimization  for Image Denoising}
\author{ Haijuan Hu, Jacques Froment, Quansheng Liu
\thanks{Haijuan Hu is with Northeastern University at Qinhuangdao,
              School of Mathematics and Statistics,  Hebei, 066004, China. She was with Univ Bretagne-Sud, CNRS UMR 6205,
           LMBA, Campus de Tohannic, F-56000 Vannes, France
              (e-mail: huhaijuan61@126.com).}
\thanks{
 Jacques Froment and Quansheng Liu are with Univ Bretagne-Sud, CNRS UMR 6205,
           LMBA, Campus de Tohannic, F-56000 Vannes, France
(e-mail: Jacques.Froment@univ-ubs.fr; Quansheng.Liu@univ-ubs.fr).}
}

\maketitle

\begin{abstract}
Patch-based sparse representation and low-rank approximation for image processing  attract much attention in recent years.
 The minimization of the matrix rank coupled with the Frobenius norm data fidelity can be solved by the hard thresholding filter with principle component analysis (PCA) or singular value decomposition (SVD). Based on this idea, we propose a patch-based low-rank minimization method for image denoising,
which learns compact dictionaries from similar patches with PCA or SVD, and applies simple hard thresholding filters to shrink the representation coefficients. Compared to recent patch-based sparse representation methods, experiments demonstrate that the proposed method is not only rather rapid, but also  effective for a variety of natural images, especially for texture parts in images.
\end{abstract}

\begin{IEEEkeywords}
  Image denoising, patch-based method,  low-rank minimization, principal component analysis, singular value decomposition, hard thresholding
\end{IEEEkeywords}

\section{Introduction} \label{chapca} 
\IEEEPARstart{I}{mage}  denoising is a classical image processing problem, but it still remains very active nowadays with the massive and easy production of digital images. We mention below some important works among the vast literature which deals with image denoising.

One  category of denoising methods concerns transform-based methods, for example \cite{donoho1994ideal,coifman1995translation}. The main idea is to calculate wavelet coefficients of images, shrink the coefficients and finally reconstruct images by inverse transform. These  methods apply fixed transform dictionaries to whole images. However, fixed dictionaries do not generally represent  whole images very well due to the complexity of natural images.   Many image details are lost while being denoised.

Another category is related to patch-based methods first proposed in \cite{buades2005review}, which explores the non-local self-similarity of natural images.  Inspired by this  ``patch-based" idea, the authors of K-SVD \cite{elad2006image} and  BM3D \cite{dabov2007image} proposed  using dictionaries to represent small local patches instead of whole images so that sparsity of coefficients can be increased,
 where the dictionaries are fixed or adaptive, and compact or overcomplete. These methods greatly improve  the traditional  methods \cite{donoho1994ideal,coifman1995translation}, leading to very good performance.
 Since these works many similar methods have been proposed to improve the denoising process, such as 
   LPG-PCA \cite{zhang2010two}, ASVD \cite{he2011adaptive}, PLOW \cite{chatterjee2012patch}, SAIST \cite{dong2013nonlocal}, NCSR \cite{dong2013nonlocally}, GLIDE \cite{talebi2014global}, and WNNM \cite{gu2014weighted}.  
However, many proposed methods are computationally complex. For example, K-SVD uses overcomplete dictionaries for sparse representation, which is time-consuming. BM3D and LPG-PCA  iterate the denoising process twice;  SAIST and WNNM iterate about 10 times. The computational cost is directly proportional to the number of iterations. 

At the same time, the low-rank matrix approximation has been widely studied and applied to image processing \cite{cai2010singular,schaeffer2013low,peng2014reweighted}. Many low-rank models have no explicit solution. However,
the paper \cite {cai2010singular} proves that the nuclear norm minimization with the Frobenius norm data fidelity can be solved by a soft thresholding filter.  (See also the paper \cite{gu2014weighted} where an alternative proof is given.) Furthermore, with the help of Eckart-Young theorem \cite{eckart1936approximation}, the paper \cite{hiriart2013eckart} demonstrates that the  solution of the exact low-rank matrix minimization problem  ($l_0$ norm)  can be obtained by a  hard thresholding filter.

Inspired by the above theories, in this paper,  a patch-based low-rank minimization (PLR) method is proposed for image denoising. First, similar patches are stacked together to construct similarity matrices. Then each  similarity matrix is denoised by minimizing  the matrix rank coupled with the Frobenius norm data fidelity. The minimizer can be obtained by a hard thresholding filter with principle component analysis (PCA) or singular value decomposion (SVD).
 The proposed method is rather rapid, since we use compact dictionaries which are more computationally efficient than over-completed dictionaries, and we do not iterate. Moreover, experiments show that the proposed method is as good as the state-of-the-art   methods, such as K-SVD \cite{elad2006image}, BM3D \cite{dabov2007image}, LPG-PCA \cite{zhang2010two}, ASVD \cite{he2011adaptive}, PLOW \cite{chatterjee2012patch}, SAIST \cite{dong2013nonlocal}, and WNNM \cite{gu2014weighted}. 


 The rest of the paper is organized as follows.
In Section II, we introduce our method.
The experimental results are shown in Section III. Finally,  this paper is concluded   in Section IV.

\section{Patch-Based Low-Rank Minimization} \label{secpc}
The  noise
model is:
$$\boldsymbol {v}=\boldsymbol{u}+\boldsymbol\eta, $$
 where $\boldsymbol u$ is the
original image, $\boldsymbol v$ is the noisy one, and $\boldsymbol \eta$ is the
Gaussian noise with mean $0$ and standard deviation
 $\sigma$. The images $\boldsymbol u,\boldsymbol v, \boldsymbol \eta$ are with size $M\times N$. Without loss of generality, we suppose that $M=N$.

\subsection{Proposed  Algorithm}
Divide the noisy image $\boldsymbol v$ into overlapped patches of size $d\times d$. Denote the set of all these patches as
$\boldsymbol {\mathcal{ S}}=\{\boldsymbol x_i: i=1,2,\cdots, (N-d+1)^2\} $. 

For each patch $\boldsymbol x \in  \boldsymbol { \mathcal S}$,  called reference patch, consider all the  overlapped patches contained in its $n\times n$ neighborhood\footnote{The reference patch is located at the center of the neighborhood, if the parities of $d$ and $n$ are the same; otherwise, the reference patch is located as near as possible to the center of the neighborhood.} (the total number of such patches is $(n-d+1)^2$ patches). Then choose the $m \;(m\geq d^2)$ most similar patches (including the reference patch itself) to the reference patch among the $(n-d+1)^2$ patches. The similarity is determined by the $l^2$-norm distance.

Next, for each reference patch, its similar  patches are reshaped as vectors, and stacked together to form a matrix of size $d^2\times m$, called similarity matrix. The similarity matrix is denoted as $\boldsymbol S=(\boldsymbol s_1, \boldsymbol s_2, \cdots, \boldsymbol s_m)$, where  columns of $\boldsymbol S$, i.e. $\boldsymbol s_i, i=1,2,\cdots, m$, are vectored similar patches.   Then all the patches in the matrix $\boldsymbol S$ are denoised together using the hard thresholding method with
  the principal component (PC) basis, or equivalently, with the singular value decomposition (SVD) basis derived from the matrix $\boldsymbol S$; the detailed process will be given afterward. For  convenience, we assume that the mean of the patches in $\boldsymbol S$, denoted by $\boldsymbol s_c:=\frac1 m \sum_{l=1}^m \boldsymbol s_l$, is 0. In practice, we subtract $\boldsymbol s_c$ from $\boldsymbol s_i$ to form the matrix $\boldsymbol S$, and add $\boldsymbol s_c$ to the final estimation $\bar {\boldsymbol s}_l$ of each patch.

  Since the patches are overlapped,  every pixel is finally estimated as the average of repeated estimates.

The process of denoising the matrix $\boldsymbol S$ is shown as follows.
Firstly, we derive adaptive basis using PCA.
The PC basis is the set of the eigenvectors of $ \boldsymbol S \boldsymbol S^T$. Write the eigenvalue decomposition\footnote{We assume that the matrix $\boldsymbol S \boldsymbol S^T$ has full rank, and it has no identical eigenvalues, which are generally true in practice.}
\begin{equation}
\boldsymbol S \boldsymbol S^T=\boldsymbol P \boldsymbol \Lambda \boldsymbol P^{-1}
\label{pcadec}
\end{equation}
with
$$\boldsymbol P=(\boldsymbol g_1,  \boldsymbol g_2, \cdots, \boldsymbol g_{d^2}), \boldsymbol \Lambda=\mbox{diag}(m\lambda_1^2,m\lambda_2^2, \cdots, m\lambda_{d^2}^2), $$
where $\boldsymbol g_i$ denotes the $i$-th column of $\boldsymbol P$ and $\mbox{diag}(c_1,c_2, \cdots) $ denotes the diagonal matrix with $(c_1,c_2, \cdots)$ on the diagonal.
The PC basis is the set of the columns of $\boldsymbol P$, that is, $\{\boldsymbol g_1,  \boldsymbol g_2, \cdots, \boldsymbol g_{d^2}\}$.

The original patches $\boldsymbol s_i$ in the similarity matrix $\boldsymbol S$ are estimated as follows: 
\begin{equation}\bar{\boldsymbol s_l}=\sum_{k=1}^{d^2} a_k\langle \boldsymbol s_l, \boldsymbol g_k \rangle \boldsymbol g_k, \quad l=1,2,\cdots,m
\label{pcwiehdv}
\end{equation}
where
\begin{equation}
 a_k=\left\{\begin{array}{rl}
1 & \mbox{if} \;\; \lambda_k^2> t^2, \\
0 & \mbox{otherwise},\end{array}\right.
\label{threshold}
\end{equation}
$t$ being the threshold.
Or equivalently,  the matrix composed of estimated patches ($\ref{pcwiehdv}$) can be written as
\begin{equation}\bar{\boldsymbol S}:=(\bar{\boldsymbol s}_1,\bar{\boldsymbol s}_2,\cdots,\bar{\boldsymbol s}_m)=\boldsymbol P h(\boldsymbol \Lambda) \boldsymbol P^{-1}\boldsymbol S,
\label{pcwiehdv2}
\end{equation}
with
\begin{equation} h(\boldsymbol \Lambda)=\mbox{diag} (a_1, a_2, \cdots, a_{d^2}).
\label{lamh}
\end{equation}
Note that 
\begin{equation}
\frac1{m}\sum_{l=1}^m (\langle \boldsymbol s_l, \boldsymbol g_k \rangle)^2=\lambda_k^2
\label{lambda}
\end{equation}
after a simple calculation. Thus $\lambda_k$ can be interpreted as the standard deviation of the basis coefficients.

We could also consider the singular value decomposition (SVD) of $ \boldsymbol S$:
\begin{equation}
 \boldsymbol S=\boldsymbol P\boldsymbol \Sigma  \boldsymbol Q^T,
 \label{svddec}
\end{equation}
where $\boldsymbol P$ is chosen as the same  orthogonal matrix in  (\ref{pcadec}),  $\boldsymbol \Sigma$ is a diagonal matrix, and $\boldsymbol Q$ (of size $m \times d^2$) has orthogonal columns such that $\boldsymbol Q^{T} \boldsymbol Q=\boldsymbol I$ with $\boldsymbol I$ the identity matrix.
Then the denoised matrix (\ref{pcwiehdv2}) is equal to
\begin{equation}
\hat{\boldsymbol S}:=\boldsymbol P H_{t\sqrt{m}}(\boldsymbol \Sigma) \boldsymbol Q^T,
\label{pcwiehdv3}
\end{equation}
where $H_{t\sqrt{m}}(\boldsymbol \Sigma)$ is a diagonal matrix, with the diagonal of $H_{t\sqrt{m}}(\boldsymbol \Sigma)$  obtained by the hard thresholding operator
\begin{equation} H_{t\sqrt{m}}(\boldsymbol \Sigma)_{kk}=\left\{\begin{array}{rl}
\boldsymbol \Sigma_{kk} & \mbox{if} \;\; \boldsymbol \Sigma_{kk}> t\sqrt{m}, \\
0 & \mbox{otherwise,}\end{array}  \quad k=1,2,\cdots, d^2. \right.
\label{hardsvd}
\end{equation}
In fact, the equality of (\ref{pcwiehdv2}) and (\ref{pcwiehdv3}) can be demonstrated as follows. By the equations (\ref{pcadec}) and (\ref{svddec}), we have $\boldsymbol \Lambda=\boldsymbol \Sigma^2$, and $\boldsymbol P^{-1}\boldsymbol S=\boldsymbol \Sigma\; \boldsymbol Q^T$. Furthermore, by the equations (\ref{lamh}) and (\ref{hardsvd}), we get $ h(\boldsymbol\Lambda) \boldsymbol \Sigma=H_{t\sqrt{m}}(\boldsymbol \Sigma)$. Thus it follows that $\bar{\boldsymbol S}=\hat{\boldsymbol S}$.
\subsection{ Low-Rank Minimization}

 Theorem \ref{theo} stated below is an unconstrained version of the Eckart-Young
  theorem \cite{eckart1936approximation}, and comes from Theorem 2(ii) in \cite{hiriart2013eckart}.  
According to Theorem \ref{theo}, it 
easily follows that
\begin{equation}
\hat{\boldsymbol S}=\arg\min_X \|\boldsymbol S-X\|_F^2+mt^2 Rank (X),
\label{pcwiehdv4}
\end{equation}
where the minimum is taken over all the matrices $X$ having the same size as $\boldsymbol S$, and $\|\cdot\|_F$ is the Frobenius norm.
   Hence the denoised matrix $\hat{\boldsymbol S}$ is the solution of  the exact low-rank minimization problem.
\begin{theo}
\label{theo}
 The following low-rank minimization  problem
\begin{equation}
\hat{X}=\arg\min_X \|Y-X\|_F^2+\mu \, Rank(X)
\label{eqori}
\end{equation}
has the  solution\footnote{Strictly speaking, if none of the singular values of $Y$ equals with $\sqrt{\mu}$, the solution is unique, which is generally true in practice. }
\begin{equation}
\hat{X}=UH_{\sqrt{\mu}}( \Sigma)V^T,
\label{xori}
\end{equation}
where $Y=U \Sigma V^T$ is the SVD of $Y$, and $H_{\sqrt{\mu}}$ is the hard thresholding operator
$$ H_{\sqrt{\mu}}( \Sigma)_{kk}=\left\{\begin{array}{rl}
 \Sigma_{kk} & \mbox{if} \;\;  \Sigma_{kk}>\sqrt{\mu}, \\
0 & \mbox{otherwise.}\end{array}\right.
$$
\end{theo}

\subsection{Choice of the Threshold}
The choice of the threshold $t$ in (\ref{threshold}) is crucial for the proposed algorithm. We study it by minimizing the mean squared error of estimated values of vectored patches $\boldsymbol s_l, (l=1,2,\cdots,m)$ in a similarity matrix $\boldsymbol S $. 
Denote $$\boldsymbol s_l=\boldsymbol u_l +\boldsymbol \eta_l,$$
where $\boldsymbol u_l$ and $\boldsymbol \eta_l$ are the vectored patches of the true image  and the noise  corresponding to $\boldsymbol s_l$ respectively.

By $(\ref{pcwiehdv})$ or (\ref{pcwiehdv2}), it can be easily obtained that
 \begin{eqnarray}
\|\bar{\boldsymbol s_l}-\boldsymbol u_l\|^2
=\sum_{k=1}^{d^2}(a_k-1)^2(\langle \boldsymbol g_k, \boldsymbol u_l \rangle)^2+\sum_{k=1}^{d^2}a_k^2(\langle \boldsymbol g_k, \boldsymbol \eta_l \rangle)^2.
\label{erri}
\end{eqnarray}
Assume that the PC basis  $\{\boldsymbol g_1,  \boldsymbol g_2, \cdots, \boldsymbol g_{d^2}\}$ only depends on the true value vectors  $\{\boldsymbol u_1,  \boldsymbol u_2, \cdots, \boldsymbol u_m\}$ and hence is independent of $\{\boldsymbol  \eta_1,   \boldsymbol \eta_2, \cdots,  \boldsymbol \eta_m\}$.
Then
\begin{equation} \mathbb{E}(\langle \boldsymbol g_k, \boldsymbol \eta_l \rangle)^2=\sigma^2.
\label{expsig}
\end{equation}
Let
\begin{equation} \theta_k^2=\frac 1 m \sum_{l=1}^{m}(\langle \boldsymbol g_k, \boldsymbol u_l \rangle)^2.
\label{thetak2}
\end{equation}
Then by (\ref{lambda}),
we obtain
\begin{equation}
  \mathbb{E}( \lambda_k^2)=\theta_k^2+\sigma^2 .
  \label{lam}
\end{equation}
Thus, from (\ref{erri}), (\ref{expsig}), and (\ref{thetak2}), it follows that
\begin{equation}
\frac 1 m \sum_{l=1}^{m} \|\bar{\boldsymbol s_l}-\boldsymbol u_l\|^2 \approx \sum_{k=1}^{d^2}(a_k-1)^2\theta_k^2+\sigma^2\sum_{k=1}^{d^2}a_k^2.
\label{erroraverage}
\end{equation}
After a simple calculation, the optimal value for $a_k$ is
$$ \hat{a}_k=\left\{\begin{array}{rl}
1 & \mbox{if} \;\; \theta_k^2> \sigma^2, \\ 
0 & \mbox{otherwise.}\end{array}\right.
$$
Since $\lambda_k^2\approx \theta_k^2+\sigma^2 $ by ($\ref{lam}$), the optimal value of the threshold in ($\ref{threshold}$) is $t^2\approx 2\sigma^2$.
In practice, we find that $t=1.5 \sigma$ is a good choice.



\section{Experimental Results}
In this section, we compare the performance of our PLR method with those of  state-of-the-art methods, including the highly competitive method WNNM \cite{gu2014weighted} proposed very recently.
Standard gray images 
%
%
%
 are utilized to test the performance of  methods. For the simulation, the level of noise is supposed to be known, otherwise there are methods to estimate it; see e.g. \cite{johnstone1997wavelet}. For each image and each level of noise, all the methods are applied to the same noisy images.

  For our algorithm, the patch size is set to $d=7$, the size of neighborhoods for selecting similar patches is set to $n=35$, and the number of similar patches in a similarity matrix is chosen as $m=5d^2$. Image boundaries are handled by assuming symmetric boundary conditions. For the sake of computational efficiency, the moving step from one reference patch to its neighbors both horizontally and vertically is  chosen as the size of patches, that is, 7.  For other comparison algorithms, we utilize the original codes released by theirs authors.

\begin{figure*}
\renewcommand{\arraystretch}{0.5} \addtolength{\tabcolsep}{-6pt} \vskip3mm {%
\fontsize{8pt}{\baselineskip}\selectfont
\begin{tabular}{ccccc}
\includegraphics[width=0.20\linewidth]{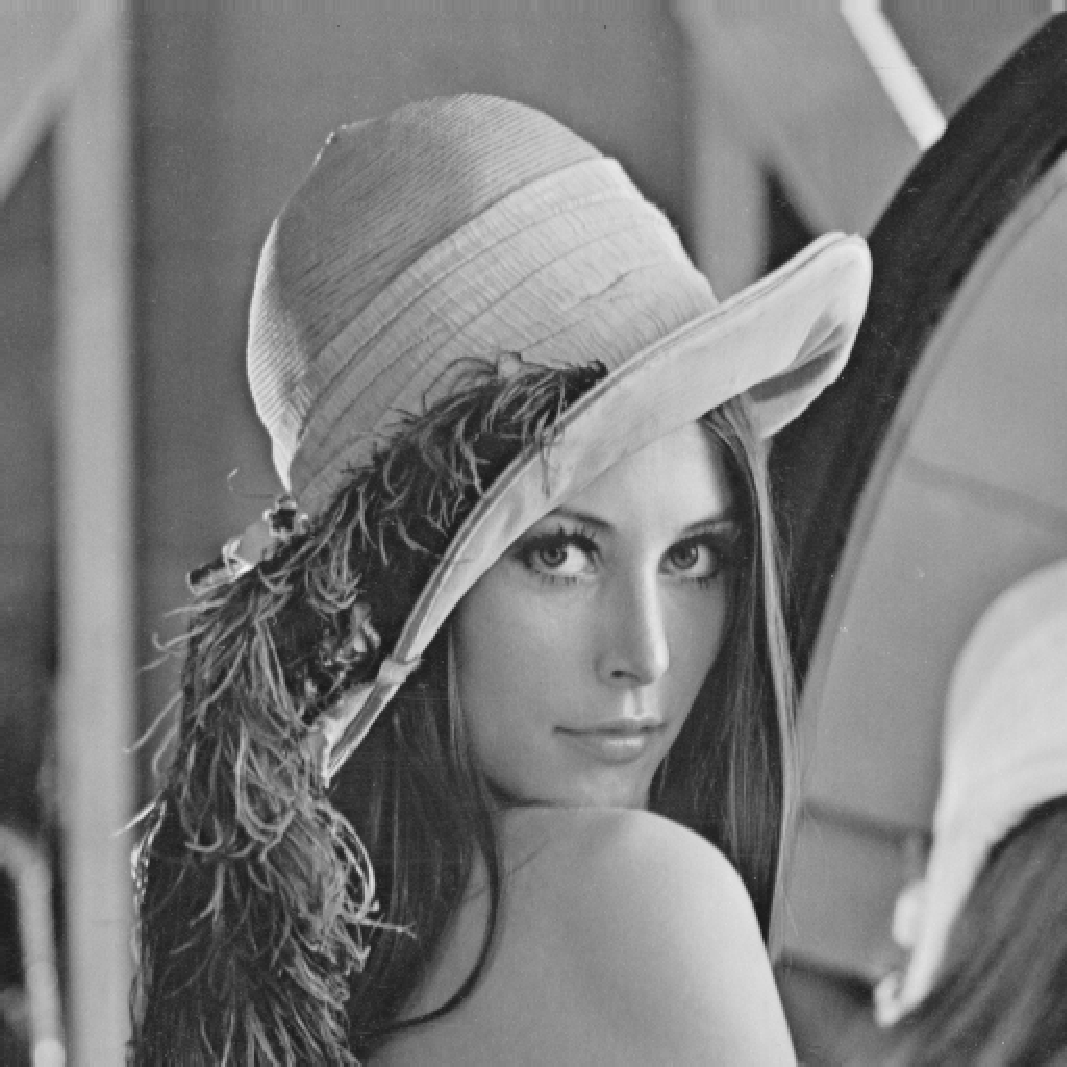}
&\includegraphics[width=0.20\linewidth]{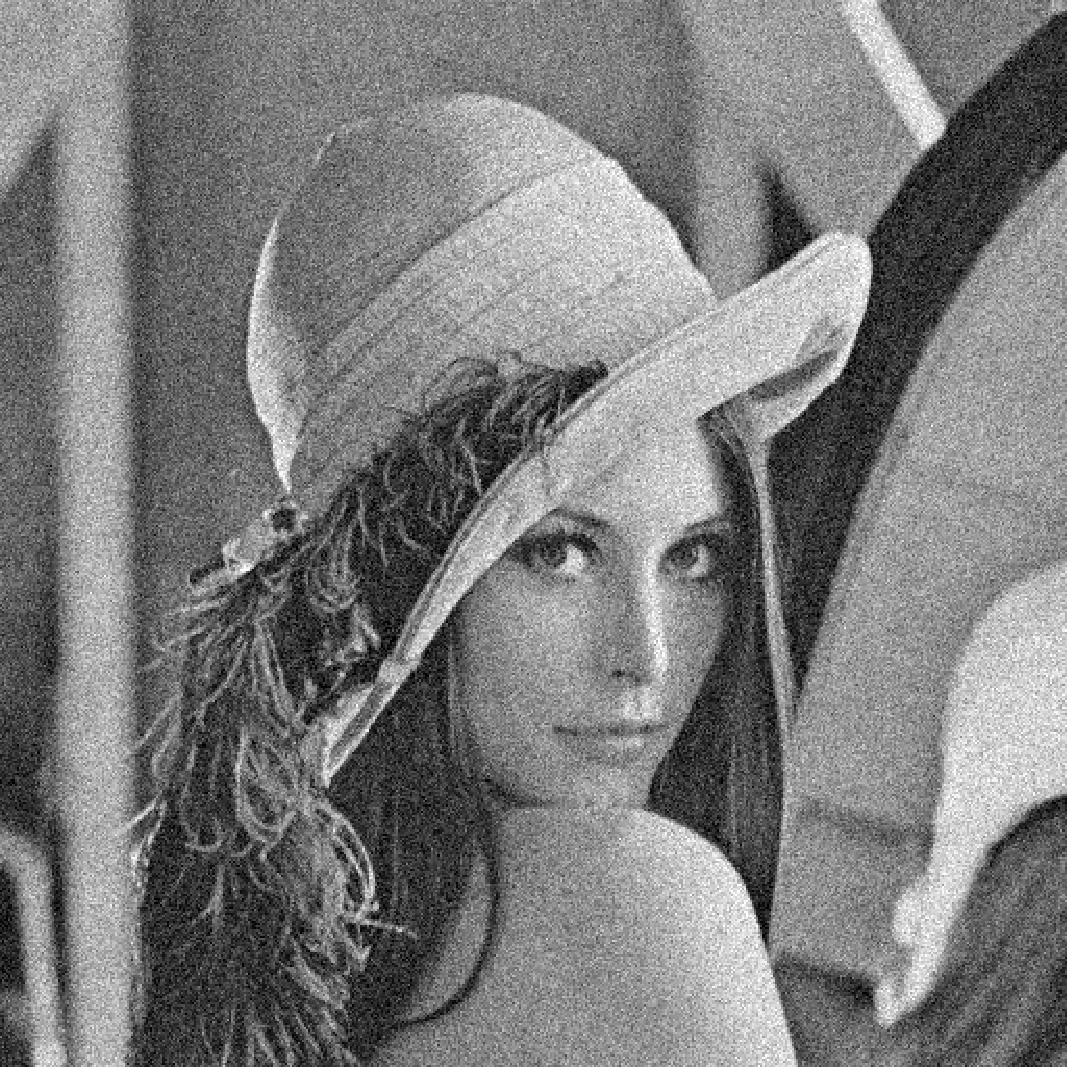}
&\includegraphics[width=0.20\linewidth]{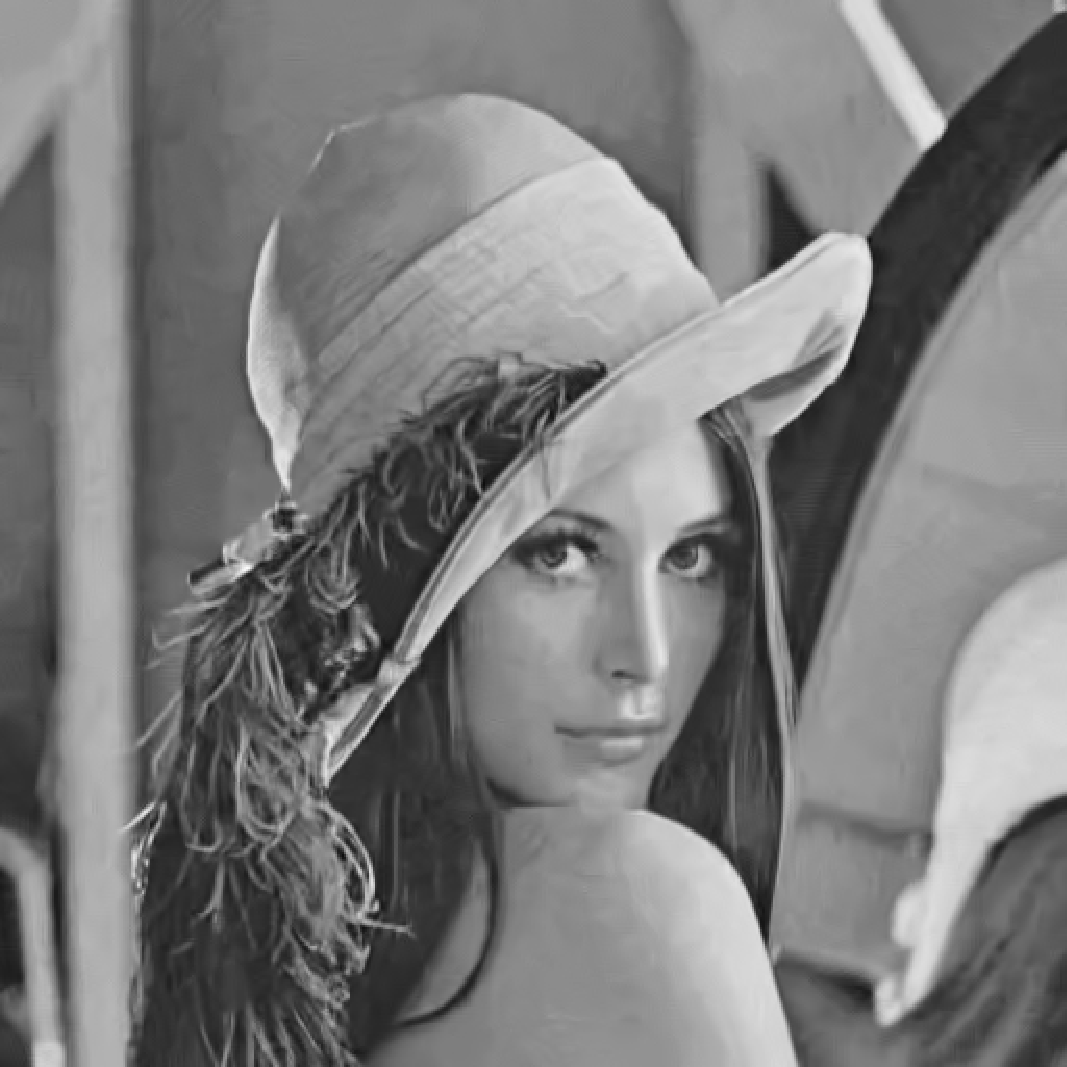}
& \includegraphics[width=0.20\linewidth]{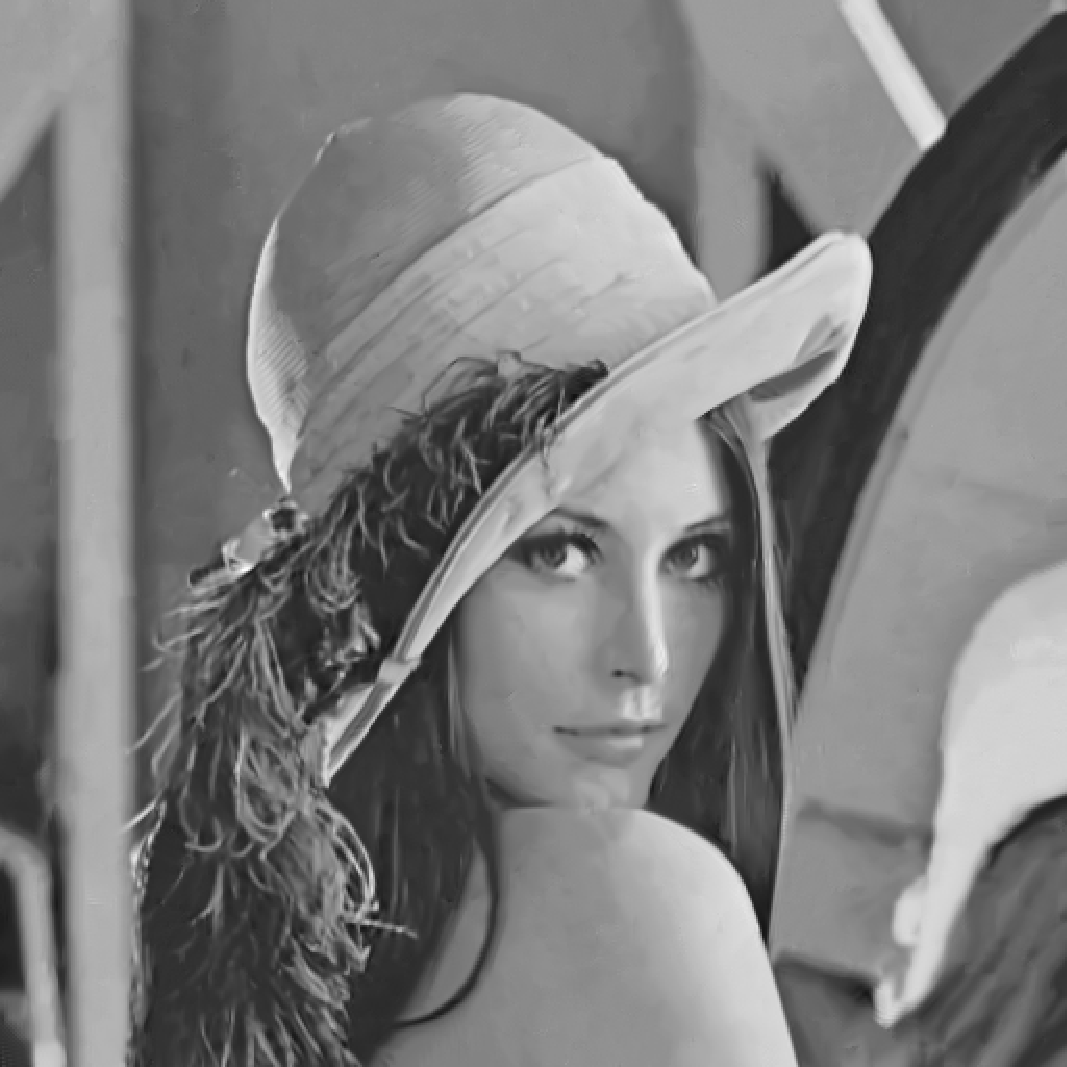}
& \includegraphics[width=0.20\linewidth]{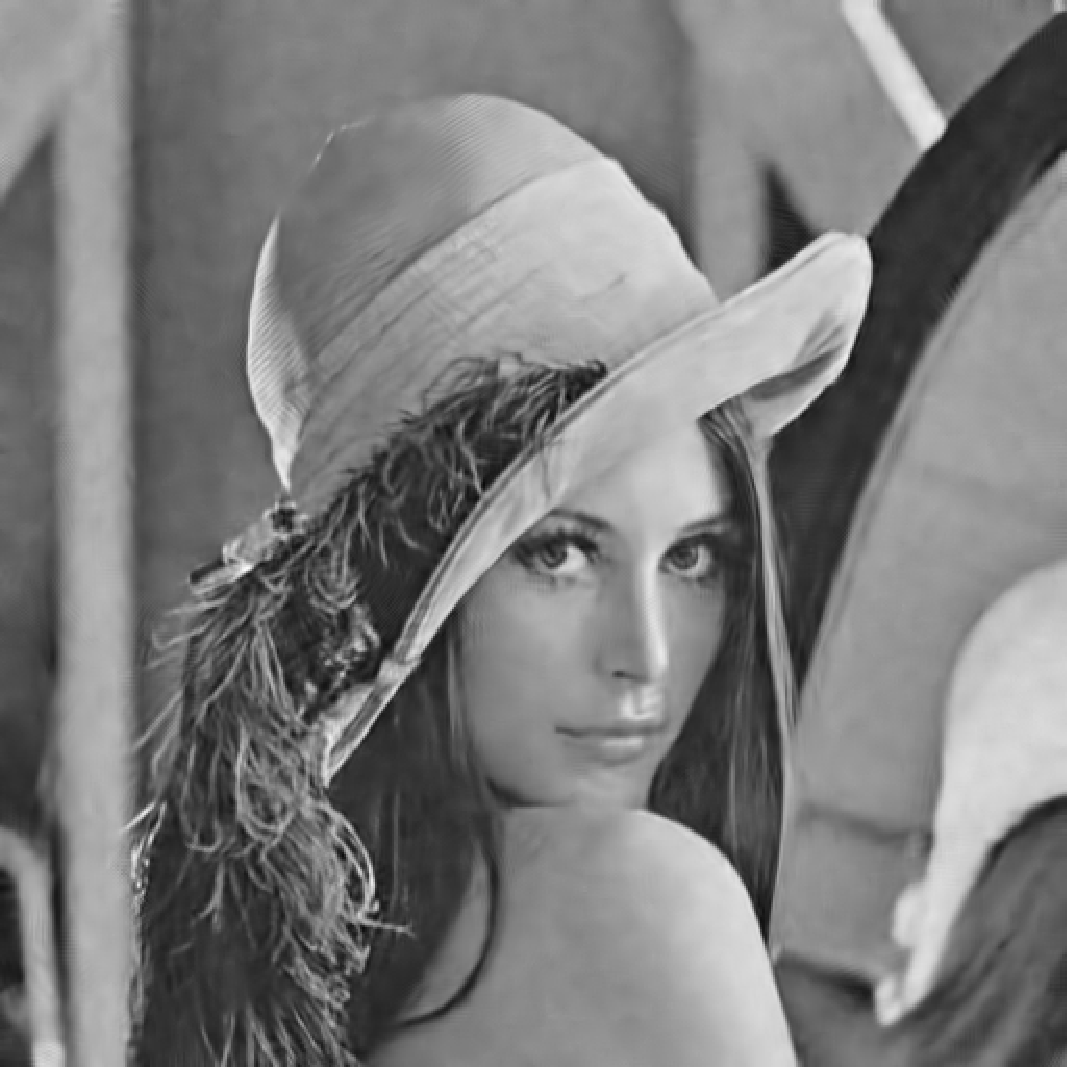}
\\ 

\includegraphics[width=0.20\linewidth]{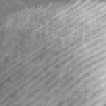}
&\includegraphics[width=0.20\linewidth]{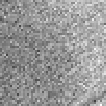}
&\includegraphics[width=0.20\linewidth]{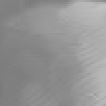}
& \includegraphics[width=0.20\linewidth]{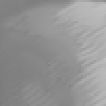}
 & \includegraphics[width=0.20\linewidth]{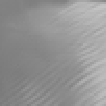}

\\
\includegraphics[width=0.20\linewidth]{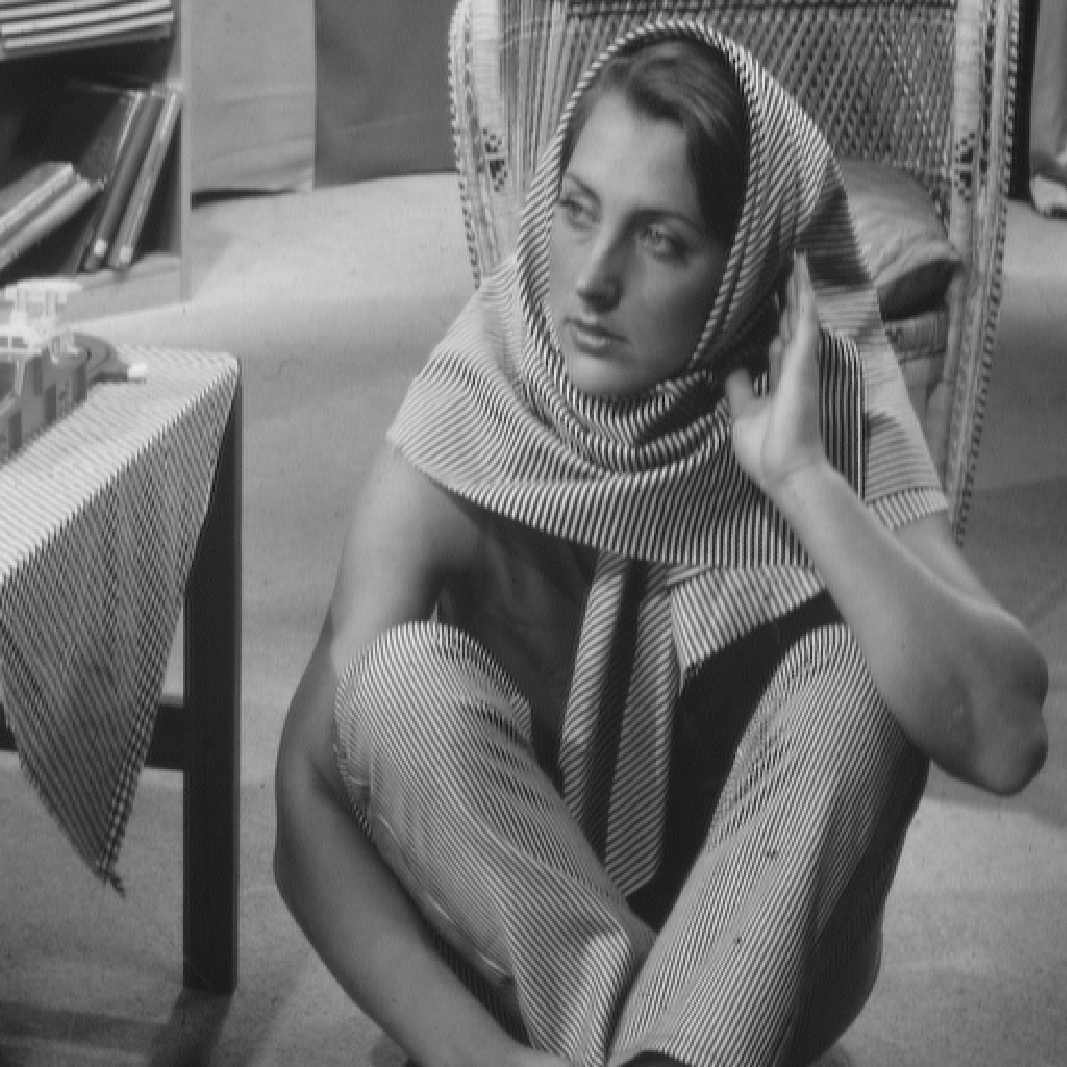}
&\includegraphics[width=0.20\linewidth]{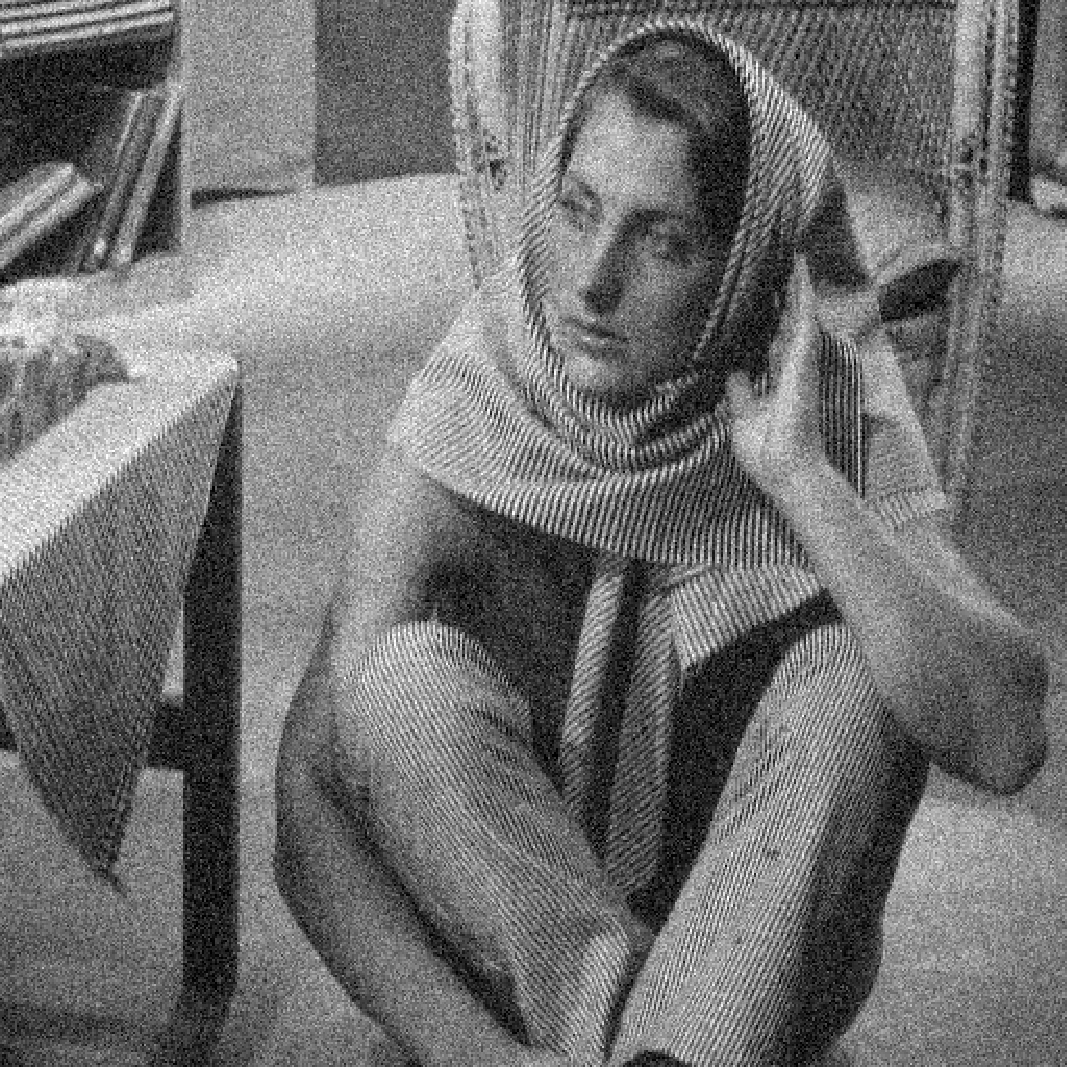}
&\includegraphics[width=0.20\linewidth]{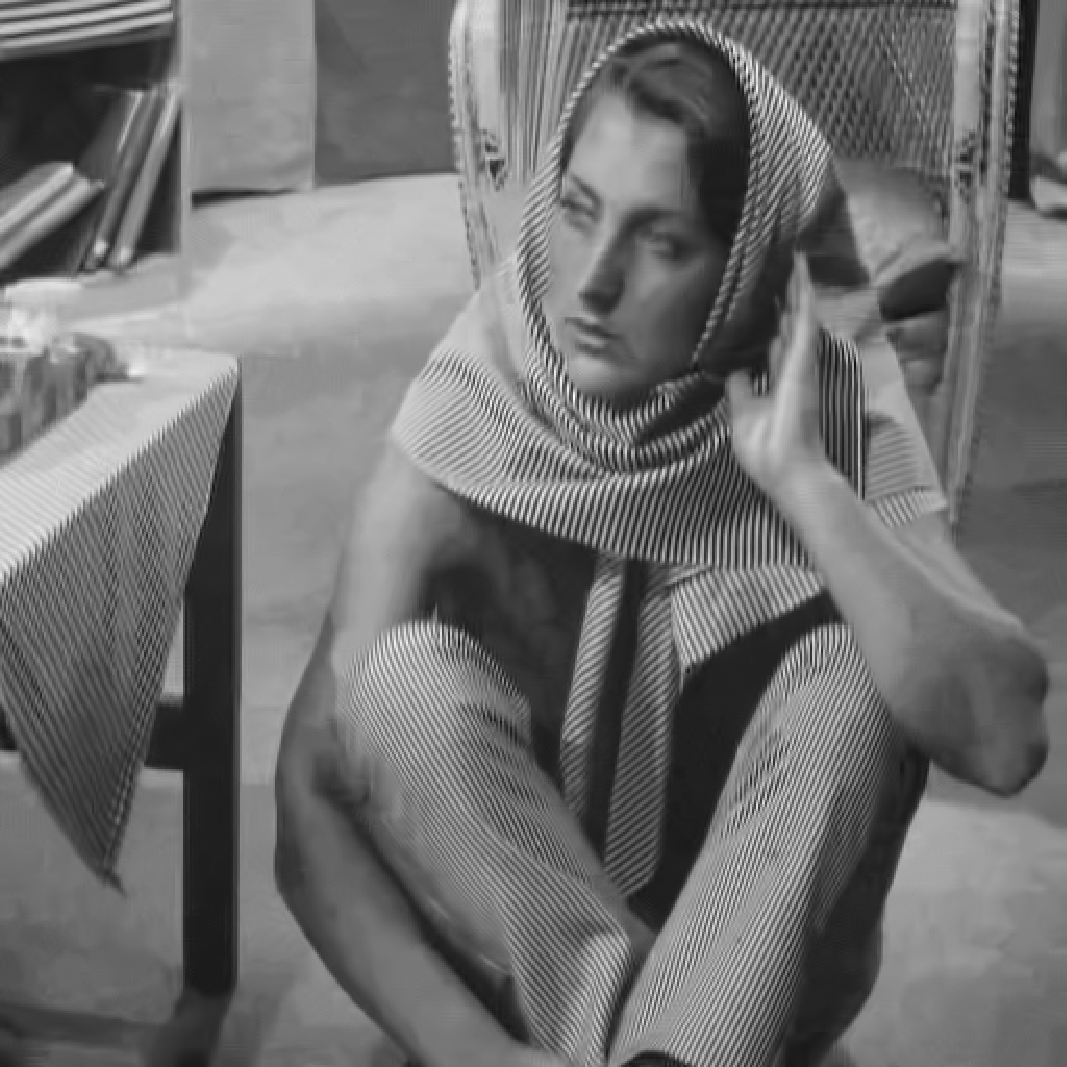}
&\includegraphics[width=0.20\linewidth]{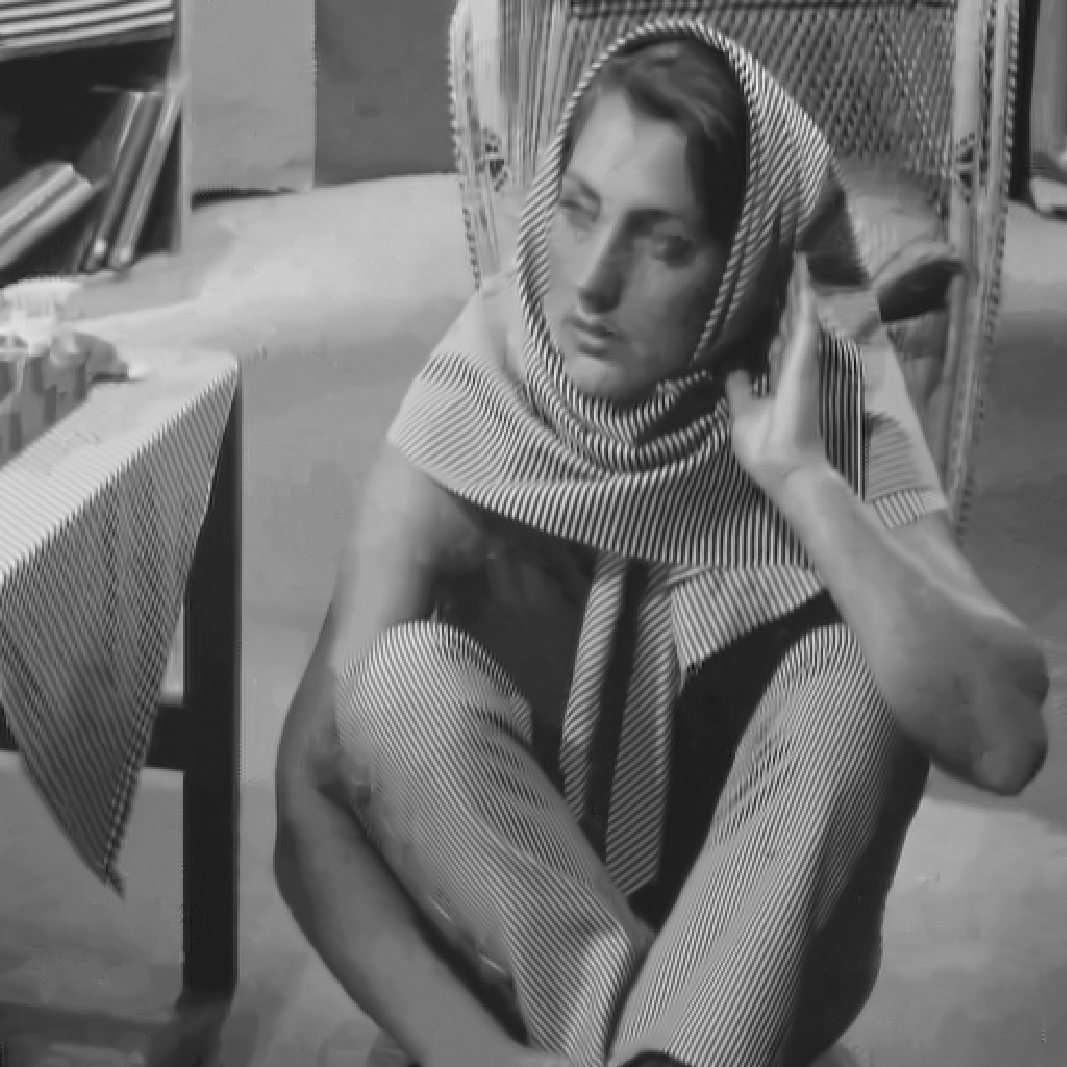}
& \includegraphics[width=0.20\linewidth]{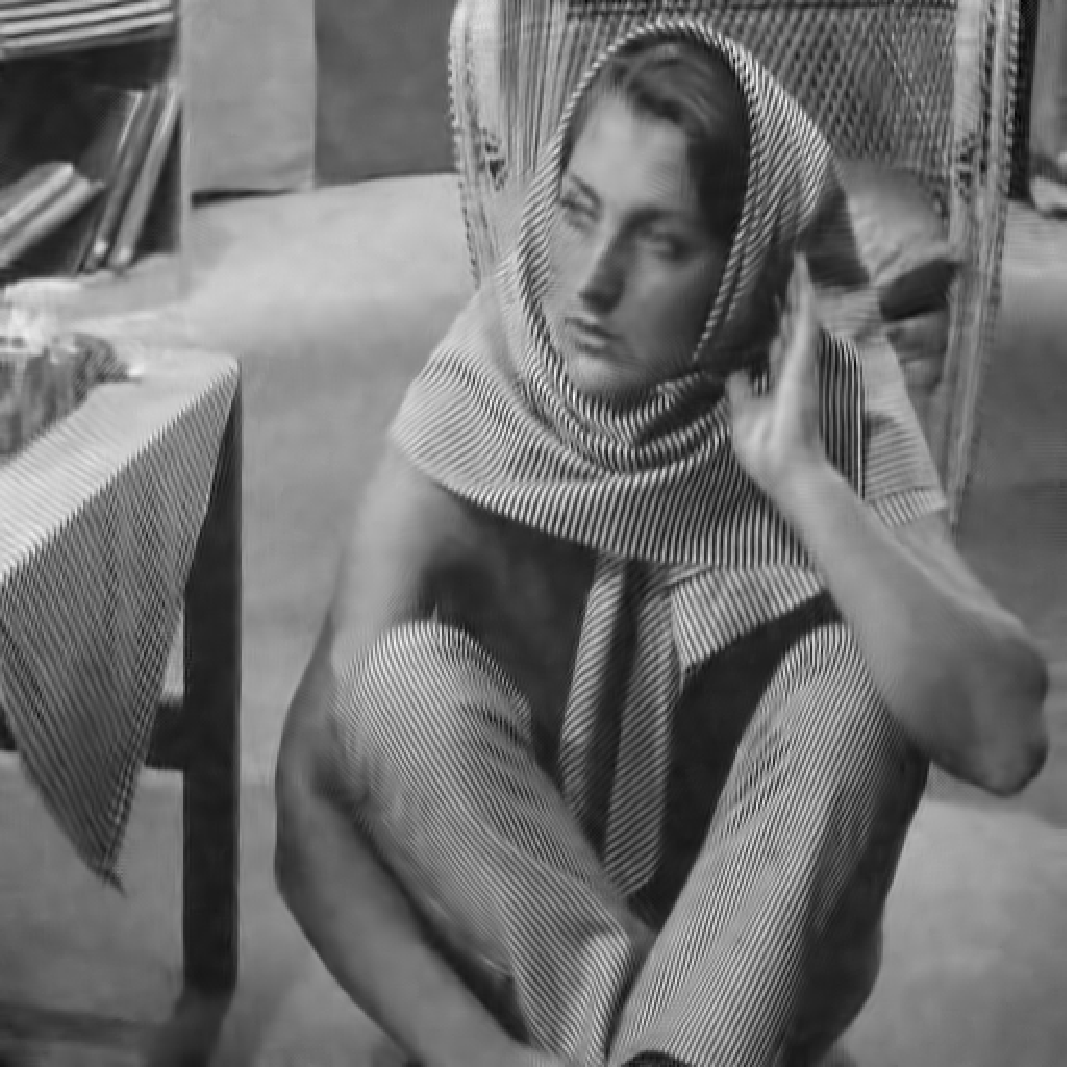}
\\
\includegraphics[width=0.20\linewidth]{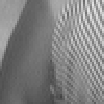}
&\includegraphics[width=0.20\linewidth]{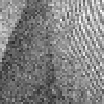}
&\includegraphics[width=0.20\linewidth]{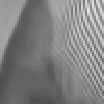}
&\includegraphics[width=0.20\linewidth]{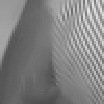}
& \includegraphics[width=0.20\linewidth]{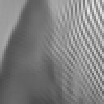}\\
Original & Noisy & BM3D\cite{dabov2007image} & WNNM \cite{gu2014weighted} & PLR \\
\end{tabular}
} \vskip1mm
\par
\rule{0pt}{-0.2pt}
\par
\vskip1mm
\caption{ Compare  denoised images Lena and Barbara by our method and other methods for $\sigma=20$. From left to right, the images are original images, noisy images, images denoised by BM3D,   WNNM, and our PLR method. To make  the differences clearer, the second row and the bottom row display parts of Lena images and Barbara images extracted from the first row and third row respectively.}
\label{pcbm1}
\end{figure*}

  In Table \ref{psnrpca}, we compare the PSNR (Peak Signal-to-Noise Ratio) values of our PLR method with other methods. The PSNR value is defined by 
  $$
  \mbox{PSNR } (\bar {\boldsymbol v}) = 20\log_{10} \frac{255N}{\|\bar {\boldsymbol v} - {\boldsymbol u}\|_F} \mbox{dB},
  $$
    where  $\boldsymbol u$ is the   original image,  and $\bar {\boldsymbol v}$ the restored one. As  can be seen in Table \ref{psnrpca}, our method is  generally better than K-SVD \cite{elad2006image}, LPG-PCA \cite{zhang2010two} and PLOW \cite{chatterjee2012patch}, and  sometimes even better than BM3D \cite{dabov2007image}. Furthermore,  for the visual comparisons, our method is also  good. For example, as can be seen  in Fig.$\ref{pcbm1}$, our method  preserves the texture parts in Lena and Barbara the best among all the methods.

To have a clear comparison of  complexities of different methods, we compare the average  CPU time to remove  noise with $\sigma=20$ for the testing images of size $256\times 256$: Peppers, House and Cameraman.  All the codes are written in M-files and run in the platform of MATLAB R2011a on a 3.40GHz Intel Core i7 CPU processor. We do not include BM3D for comparison since the original code of BM3D contains MEX-files. The running time is displayed in second in Table \ref{runtm}. The comparisons clearly show that the  proposed method is much faster than the others.

\begin{table}
\begin{center}
\caption{PSNR values for removing  noise for our  PLR and other methods. Cam is the Cameramen image}
{\footnotesize\addtolength{\tabcolsep}{-3pt}
\begin{tabular}{|c||c|c|c|c|c|c|c|c|c} 
\hline
Image  &Lena &Barbara& Peppers &Boats &Bridge & House & Cam \\
\hline
\multicolumn{8}{|c|}{$\sigma=10$}\\
\hline 
 K-SVD\cite{elad2006image} & 35.50 & 34.82 & 34.23& {33.62}& 30.91 & {35.96} & 33.74 \\
\hline
LPGPCA\cite{zhang2010two} &35.72&35.46&34.05 &  33.61 &   30.86 & 36.16  &  33.69 \\
\hline
ASVD\cite{he2011adaptive} &35.58&  35.58&   33.55& 33.26&  27.76&   36.46&  31.62\\
\hline
PLOW\cite{chatterjee2012patch} & 35.29 &34.52 &33.56 &32.94 & 29.88 &36.22 & 33.15 \\
\hline
 \textbf{PLR} & ${35.90}$ &  35.50  & 34.28  & 33.76  & 30.78 &  36.57  & 33.73\\
\hline
 BM3D\cite{dabov2007image}& 35.90 &  ${35.39}$ &  ${ 34.68}$&   ${33.88}$ &   ${31.06}$ &   ${36.71}$ &   ${34.18}$ 
\\
\hline
 SAIST\cite{dong2013nonlocal} &35.87 &35.69 &34.76 &33.87 &31.03 &36.52 &34.28 \\
\hline

WNNM\cite{gu2014weighted}  & 36.02& 35.92&   34.94&   34.05&   31.16&   36.94&   34.44\\
\hline
\multicolumn{8}{|c|}{$\sigma=20$}\\
\hline
K-SVD\cite{elad2006image} & 32.38&{31.12}&{30.78}&30.37& 27.03&{33.07}&{30.01}\\
\hline
LPGPCA\cite{zhang2010two} & 32.61 & 31.69 & 30.50 & 30.26& 26.84 & 33.10 & 29.77 \\
\hline
ASVD\cite{he2011adaptive} & 33.21&   32.96&   30.56&   31.79&   25.51&   33.53&  29.33\\
\hline
PLOW\cite{chatterjee2012patch} & 32.70 &31.48 & 30.52 & 30.36 & 26.56 & 33.56 &29.59 \\
\hline
\textbf{PLR} &33.03&  32.12&   30.90&  30.64&   27.20&   33.36&   30.12\\
\hline
 BM3D\cite{dabov2007image} &33.03 &   32.07 &  ${ 31.28}$ &   ${30.85}$ &   27.14&  ${ 33.77}$ &   ${30.48}$ \\
\hline
SAIST\cite{dong2013nonlocal}& 33.07 &32.43 &31.28 &30.78 &27.20 &33.80 &30.40 \\
\hline
WNNM\cite{gu2014weighted} & 33.10 &  32.49  & 31.53  & 30.98  & 27.29&   34.01 & 30.75 \\

\hline
\end{tabular}
}

\label{psnrpca}
\end{center}
\end{table}

\begin{table}
\begin{center}
\caption{Running time in second for our  PLR and other methods to remove  noise with images of size $256\times 256$}
{\footnotesize\addtolength{\tabcolsep}{-3pt}
\begin{tabular}{|c|c|c|c|c|c|c|ccc} 
\hline
K-SVD & LPG-PCA &ASVD & PLOW& \textbf{PLR} & SAIST &WNNM\\
\hline
210 &138  &337 & 43 & 2& 25 & 134\\
\hline
\end{tabular}
}

\label{runtm}
\end{center}
\end{table}

\section{Conclusion}
In this paper,  a patch-based low-rank minimization method for image denoising is proposed, which stacks similar patches into  similarity matrices, and denoises each  similarity matrix by seeking the minimizer of  the matrix rank  coupled with  the Frobenius norm  data fidelity. The minimizer can be obtained by a hard threshoding filter with principle component basis or  left singular vectors. The proposed method is not only rapid, but also effective compared to recently reported methods. 



\end{document}